\documentclass[conference]{IEEEtran}
\IEEEoverridecommandlockouts
\usepackage{cite}
\usepackage{amsmath,amssymb,amsfonts}
\usepackage{algorithmic}
\usepackage{graphicx}
\usepackage{textcomp}
\usepackage{xcolor}
\definecolor{cvprblue}{rgb}{0.21,0.49,0.74}
\usepackage[
  colorlinks=true,
  linkcolor=black,
  citecolor=black,
  urlcolor=cvprblue
]{hyperref}
\usepackage{booktabs}
\usepackage{multirow} 
 \usepackage{url} 
\def\BibTeX{{\rm B\kern-.05em{\sc i\kern-.025em b}\kern-.08em
    T\kern-.1667em\lower.7ex\hbox{E}\kern-.125emX}}
\begin{document}

\title{GlobalPaint: Spatiotemporal Coherent Video Outpainting with Global Feature Guidance}

\author{
Yueming Pan\textsuperscript{1,2  *},
Ruoyu Feng\textsuperscript{3},
Jianmin Bao\textsuperscript{2},
Chong Luo\textsuperscript{2 †},
Nanning Zheng\textsuperscript{1 †}\\[4pt]
\textsuperscript{1}State Key Laboratory of Human-Machine Hybrid Augmented Intelligence,\\
Institute of Artificial Intelligence and Robotics, Xi’an Jiaotong University\\[2pt]
\textsuperscript{2}Microsoft Research Asia \quad
\textsuperscript{3}University of Science and Technology of China
}

\maketitle

\begingroup
\renewcommand\thefootnote{\fnsymbol{footnote}}
\setcounter{footnote}{0}
\footnotetext[1]{This work was performed during Yueming Pan's internship at MSRA.}
\footnotetext[2]{Corresponding author.}
\endgroup

\begin{abstract}

Video outpainting extends a video beyond its original boundaries by synthesizing missing border content. Compared with image outpainting, it requires not only per-frame spatial plausibility but also long-range temporal coherence, especially when outpainted content becomes visible across time under camera or object motion. We propose \textit{GlobalPaint}, a diffusion-based framework for spatiotemporal coherent video outpainting. Our approach adopts a hierarchical pipeline that first outpaints key frames and then completes intermediate frames via an interpolation model conditioned on the completed boundaries, reducing error accumulation in sequential processing. At the model level, we augment a pretrained image inpainting backbone with (i) an Enhanced Spatial-Temporal module featuring 3D windowed attention for stronger spatiotemporal interaction, and (ii) global feature guidance that distills OpenCLIP features from observed regions across all frames into compact global tokens using a dedicated extractor. Comprehensive evaluations on benchmark datasets demonstrate improved reconstruction quality and more natural motion compared to prior methods. Our demo page is \url{https://yuemingpan.github.io/GlobalPaint/}

\end{abstract}



\section{Introduction}
\begin{figure*}[!t]
  \centering
  \includegraphics[width=\textwidth]{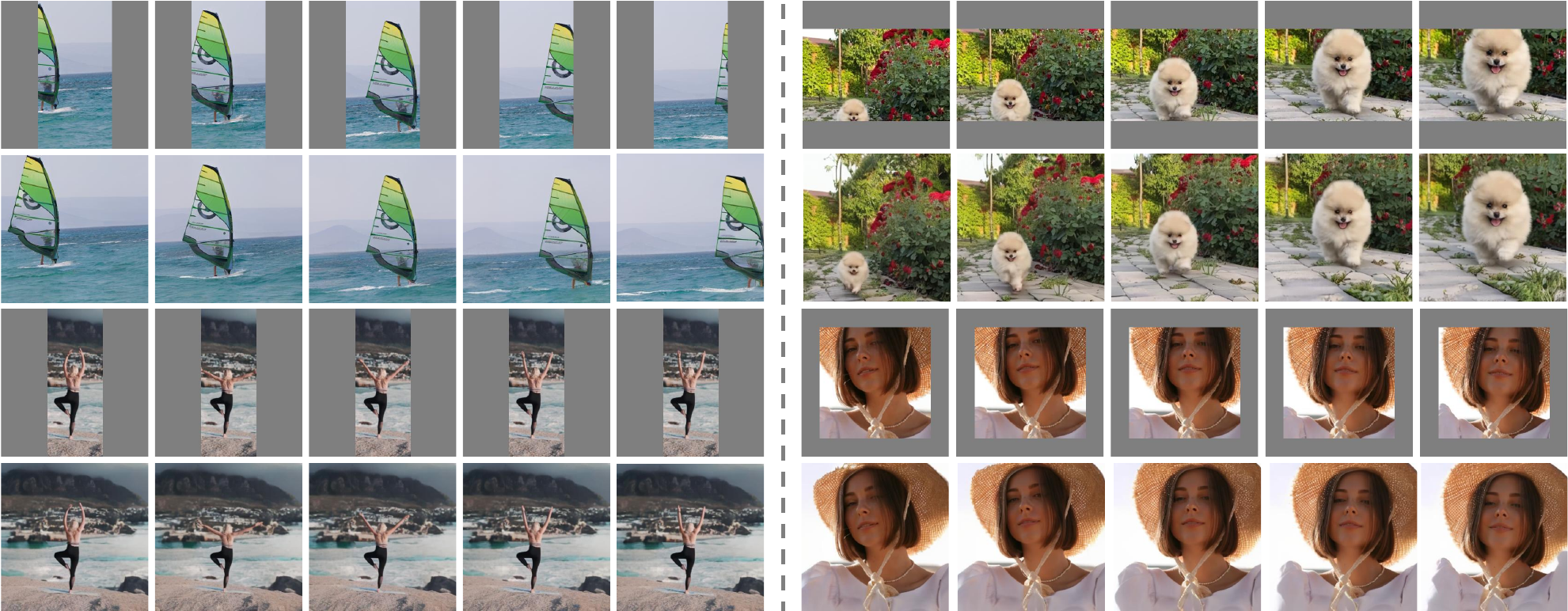}
  \caption{GlobalPaint is a diffusion-based video outpainting framework equipped with enhanced spatiotemporal modules and guided by global features. It generates spatiotemporally coherent results under various mask patterns.}
  \label{fig:teaser}
\end{figure*}

Video outpainting aims to extend a video beyond its original boundaries by synthesizing visually plausible content along the missing borders. It enables a range of multimedia applications, including aspect-ratio adaptation, shot reframing, and post-production editing where missing regions must be completed seamlessly. Despite rapid progress in image inpainting and outpainting~\cite{RenderDiffusion, Palette, BlendedDiffusion, controlnet} , video outpainting remains challenging because it requires not only per-frame spatial plausibility but also long-range temporal coherence across the entire sequence.
This challenge becomes pronounced under camera or object motion: missing regions may correspond to different parts of the scene across frames, and content synthesized in one frame can later become visible (or has already been visible) in other frames. For example, in Fig.~\ref{fig:teaser}, the sail needs to be outpainted in the final frame, while its appearance has been observed earlier in the sequence. Producing a consistent sail therefore requires leveraging global information beyond local spatiotemporal neighborhoods. Addressing this issue requires improvements at both the approach level for processing videos under limited temporal context and the model level for integrating global and spatiotemporal cues into generation.

Traditional video outpainting methods often rely on optical flow for motion modeling~\cite{Dehan}. While effective for background filling or rigid objects under mild motion, they are limited when scenes exhibit complex dynamics or significant appearance changes. Diffusion models~\cite{DDPM,DDIM} have recently shown strong generation capability in image and video synthesis~\cite{LDM,controlnet}, making them an appealing foundation for video outpainting. However, diffusion models are computationally expensive and typically operate on short clips (e.g., a few to a few dozen frames). Yet even a short 3-second video may contain around 90 frames, making joint processing of the full sequence impractical.

To address the tension between limited context length and the need for global coherence, we adopt a hierarchical processing strategy. We first select key frames from the full sequence and outpaint them to establish globally consistent content. We then outpaint the intermediate frames using an interpolation model conditioned on the completed boundary key frames, which substantially reduces error accumulation and mitigates temporal inconsistencies that often arise from naive sequential clip-by-clip processing.

At the model level, we present \textit{GlobalPaint}, a diffusion-based video outpainting model designed to achieve spatiotemporal coherence with global awareness. Built upon a text-guided image inpainting backbone, GlobalPaint introduces two key designs. First, beyond standard 1D temporal modules commonly used to adapt image models to video~\cite{AnimateDiff, videoLDM}, we incorporate 3D windowed attention to better capture joint spatiotemporal interactions. This design is particularly beneficial under our hierarchical pipeline: unlike sequential processing that operates on short consecutive clips with limited global context, key-frame processing exposes richer long-range scene information, and the enlarged spatiotemporal receptive field enables stronger interaction of such global cues, yielding more reliable key-frame completions for subsequent interpolation. Second, to explicitly leverage global cues across the entire sequence at manageable cost, we extract OpenCLIP~\cite{openclip} features from the observed regions of all input frames and condense them into a compact set of global tokens via a dedicated global feature extractor, which are injected into the denoising process through cross-attention.

We evaluate GlobalPaint on two widely used benchmarks, DAVIS~\cite{DAVIS2016} and YouTube-VOS~\cite{youtubevos}. Experiments show that GlobalPaint improves reconstruction quality and yields more natural motion. On DAVIS, GlobalPaint achieves an FVD of 227.8, reducing FVD by 24.1\% and 20.4\% compared to M3DDM~\cite{M3DDM} and MOTIA~\cite{MOTIA}, respectively. On YouTube-VOS, GlobalPaint consistently improves PSNR/SSIM/LPIPS over prior methods and attains a competitive FVD.

In summary, our contributions are three-fold:
\begin{itemize}
  \item We propose a hierarchical video outpainting framework that improves long-range consistency under the limited temporal context of diffusion models.
  \item We introduce GlobalPaint, a diffusion-based model with enhanced spatiotemporal modules and global feature guidance for coherent video outpainting.
  \item We conduct extensive quantitative and qualitative evaluations, demonstrating consistent improvements over existing approaches.
\end{itemize}

\section{Related Work}

\subsection{Diffusion-based Image Inpainting and Outpainting}
Diffusion models (DMs)~\cite{RenderDiffusion, Palette, BlendedDiffusion, controlnet} have achieved strong performance on image inpainting and outpainting. Representative approaches include latent-space inpainting with masked-image inputs~\cite{LDM}, conditioning via auxiliary branches~\cite{controlnet}, and latent blending for text-guided completion~\cite{BlendedLatentDiffusion}. These methods establish strong image-level priors, whereas our work extends diffusion-based completion to video outpainting, where temporal coherence and long-range consistency are critical.

\subsection{Video Inpainting}
Video inpainting aims to fill missing regions while preserving temporal consistency. Prior works largely follow two paradigms: flow-guided propagation pipelines~\cite{flow1,flow2,flow3,flow4,E2FGVI} and transformer-based spatiotemporal feature propagation~\cite{Fuseformer,Transformer_inpaint1,Decoupled_Spatial_Temporal_Transformer,ProPainter}. More recently, diffusion models have been adapted to video inpainting by incorporating temporal modules or flow-based latent propagation~\cite{AVID, FGDVI}. Despite relatedness, video outpainting differs from inpainting in that it extends video boundaries from sparse edge observations and often requires synthesizing substantially larger regions, making global temporal reasoning more critical.

\subsection{Video Outpainting}
For video outpainting, Dehan \emph{et al.}~\cite{Dehan} use optical-flow-based background warping from adjacent frames followed by image completion, which can struggle under large camera motion or complex dynamics. M3DDM~\cite{M3DDM} adapts an image diffusion model to videos via 1D temporal layers with mask modeling, yet maintaining coherent motion and sharp boundaries in dynamic scenes remains challenging. MOTIA~\cite{MOTIA} introduces input-specific adaptation to capture video-specific patterns, but its effectiveness depends on the richness of the available input content and relies on per-sample fine-tuning, which increases inference latency. In contrast, our method strengthens spatiotemporal modeling and global-context utilization through 3D windowed attention and compact global feature guidance, leading to more coherent and robust video outpainting.

\begin{figure*}[!t]
  \centering
  \includegraphics[width=\textwidth]{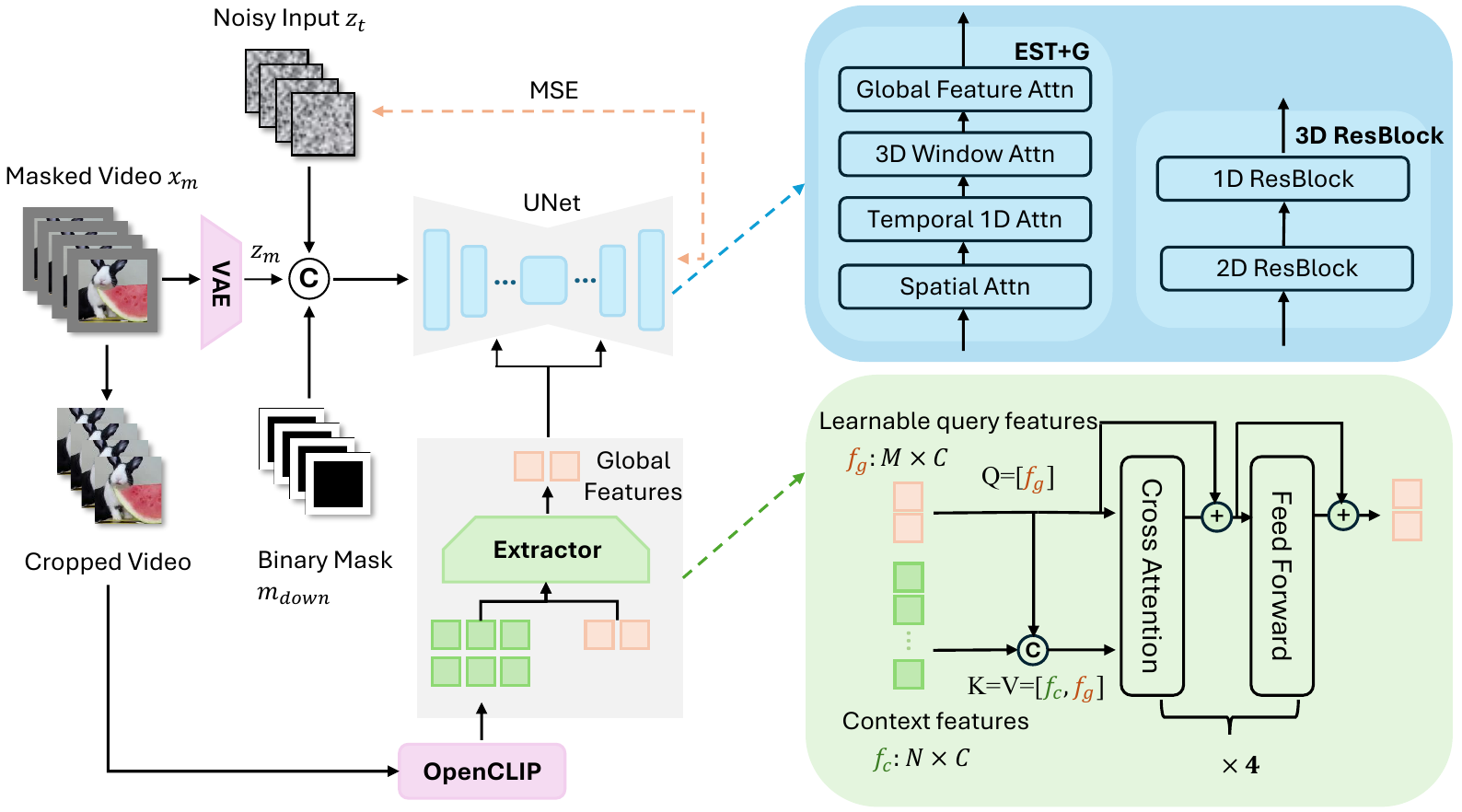}
  \caption{Overview of GlobalPaint. We follow the LDM inpainting formulation~\cite{LDM} by concatenating the noisy latent $z_t$, masked latent $z_m$, and downsampled mask $m_{\text{down}}$ as inputs to an inflated UNet.
Global cues are extracted from the observed regions of all frames using a frozen OpenCLIP encoder~\cite{openclip}, distilled into compact global tokens by a dedicated Global Feature Extractor, and injected into the UNet to guide generation process. \textbf{EST+G} denotes the Enhanced Spatial-Temporal module with Global Feature guidance.}
  \label{framework}
\end{figure*}

\section{Method}
\subsection{Preliminaries}
We build upon Latent Diffusion Models (LDMs)~\cite{LDM}, which perform diffusion in a compact latent space encoded by a variational autoencoder (VAE)~\cite{kingma2013auto}. Given an input image $x_0$, the VAE encoder $\mathcal{E}$ maps it to a latent $z_0=\mathcal{E}(x_0)$. A forward noising process produces $z_t$ by adding Gaussian noise to $z_0$ at diffusion step $t$. Conditioned on the text embedding $c=\tau(y)$ from a text encoder $\tau$, the denoising network $\epsilon_\theta$ is trained to predict the injected noise:
\begin{equation}
\mathcal{L}_{\text{LDM}}=\mathbb{E}_{t,\epsilon}\!\left[\left\| \epsilon-\epsilon_\theta(z_t,t,c)\right\|_2^2\right],
\end{equation}
where $\epsilon\sim\mathcal{N}(0,I)$.

We further adopt the LDM inpainting formulation~\cite{LDM}. Let $m\in\{0,1\}^{H\times W}$ be a binary mask where $m=0$ denotes known regions and $m=1$ denotes regions to be synthesized. The masked input is encoded as
$z_m=\mathcal{E}\!\left(x_0\odot(1-m)\right)$, and the mask is downsampled to the latent resolution as $m_{\text{down}}$.
The inpainting objective is
\begin{equation}
\mathcal{L}_{\text{inp}}=\mathbb{E}_{t,\epsilon}\!\left[\left\| \epsilon-\epsilon_\theta(z_t,t,c,z_m,m_{\text{down}})\right\|_2^2\right].
\end{equation}

\subsection{Overview of the Framework}
GlobalPaint adopts a hierarchical pipeline for video outpainting. We first sample key frames at a fixed interval and outpaint them using the base outpainting model. We then outpaint the intermediate frames between each key-frame pair with an interpolation model conditioned on the completed boundary frames. The interpolation model shares the same architecture as the base model and is obtained via brief fine-tuning on full-frame-rate clips. 
Compared with M3DDM~\cite{M3DDM}, which caters to various frame rates with a single model, our approach adopts a streamlined key-frame-first design that factorizes video outpainting into key-frame completion and interpolation. This design simplifies the model learning difficulty and reduces training cost, while delivering competitive performance.

\subsection{The GlobalPaint Model}
As shown in Fig.~\ref{framework}, we follow the LDM inpainting formulation. During training, we concatenate the noisy latent $z_t$, masked latent $z_m$, and downsampled mask $m_{\text{down}}$ along the channel dimension and feed them into an inflated UNet backbone initialized from Stable Diffusion inpainting model~\cite{LDM}.

\noindent\textbf{Enhanced Spatial-Temporal (EST) Module. }
Prior works~\cite{AnimateDiff, videoLDM} typically adapt image diffusion models to video by adding 1D temporal convolutions and attention layers. In contrast, we augment the backbone with EST modules that extend the receptive field jointly over spatial and temporal dimensions by integrating 3D windowed attention. This design is particularly effective in our hierarchical setting: key-frame processing exposes richer long-range scene context than short consecutive clips, and the expanded spatiotemporal receptive field better integrates long-range global cues, producing more reliable key frames for subsequent interpolation.

Specifically, each EST block consists of a 1D temporal block followed by a 3D windowed attention block, with zero-initialized output projections for stable training. Formally, we denote $\mathcal{T}_{1\mathrm{D}}(\cdot)$ as the 1D temporal block,
$\mathcal{T}_{\mathrm{W}}(\cdot)$ as the 3D windowed attention block,
and $\mathcal{Z}_{1\mathrm{D}}(\cdot)$, $\mathcal{Z}_{\mathrm{W}}(\cdot)$ as
zero-initialized projection layers.
\begin{equation}
\mathrm{EST}(\cdot)
=
\mathcal{Z}_{\mathrm{W}}
\!\left(
\mathcal{T}_{\mathrm{W}}
\!\left(
\mathcal{Z}_{1\mathrm{D}}
\!\left(
\mathcal{T}_{1\mathrm{D}}(\cdot)
\right)
\right)
\right).
\end{equation}
Given a spatial block $\mathcal{T}_{\mathrm{S}}(\cdot)$,
the overall update from input feature $u$ to output feature $v$ is written as
\begin{equation}
v = \mathrm{EST}\!\left(\mathcal{T}_{\mathrm{S}}(u)\right),
\end{equation}
where $u,v \in \mathbb{R}^{C\times T\times H\times W}$,
and $C$, $T$, $H$, $W$ denote the number of channels, frames, height, and width,
respectively.
During training, we freeze the spatial layers to preserve the image inpainting prior and only optimize the newly added temporal modules.

\noindent\textbf{Global Feature Guidance. }
While the EST module enlarges the spatiotemporal receptive field, it still cannot capture all information across the entire video sequence. A straightforward solution is full attention, but it incurs prohibitive computation. To incorporate long-range cues at manageable cost, we introduce compact global features distilled from all frames to guide generation.

Specifically, as shown in Fig.~\ref{framework}, we use a frozen OpenCLIP image encoder~\cite{openclip} to extract $N$ context tokens from the observed regions of all cropped frames. Since these tokens are highly redundant, directly attending to all of them is expensive. Inspired by Perceiver-style token resampling~\cite{perceiver,flamingo,openflamingo,ip_adapter}, we maintain a small set of $M$ learnable query tokens ($M\!\ll\!N$), and update them by cross-attending to the context tokens to obtain compact global representations.

The Global Feature Extractor (right-bottom of Fig.~\ref{framework}) stacks cross-attention and a feed-forward network with residual connections. Formally, given query tokens $f_g$ and context tokens $f_c$, we apply standard attention with:
\begin{equation}
Q=W^{Q}f_g,\quad K=W^{K}[f_c,f_g],\quad V=W^{V}[f_c,f_g],
\end{equation}
where $[\cdot]$ indicate the concatenation operation, $W^Q$, $W^K$, $W^V$ are learnable matrices that map inputs to query, key, and value tokens. The final global tokens $g$ are injected into the UNet through cross-attention to guide the outpainting process.

\noindent\textbf{Training Objective. }
With global tokens $g$, our final objective is
\begin{equation}
\label{eq:training}
\mathcal{L}_{\text{GP}} =
\mathbb{E}_{t,\epsilon}\!\left[\left\| \epsilon - \epsilon_{\theta}(z_t, t, c, z_m, m_{\text{down}}, g) \right\|_2^2\right].
\end{equation}

\noindent\textbf{Interpolation Model. }
To outpaint intermediate frames between two completed key frames, we fine-tune an interpolation model from the base outpainting model. Different from interpolation settings that condition only on boundary key frames~\cite{videoLDM}, we additionally provide masked intermediate frames as inputs, which supplies extra observed evidence and simplifies the completion task.

During training, we sample short clips where the first and last frames are provided as complete ground truth, and the intermediate frames are partially masked. The loss extends Eq.~\ref{eq:training} by conditioning on the boundary latents:
\begin{equation}
\mathcal{L}_{\text{interp}} =
\mathbb{E}_{t,\epsilon}\!\left[\left\| \epsilon - \epsilon_{\theta}(z_t, t, c, z_m, m_{\text{down}}, g, z_{\text{first}}, z_{\text{last}}) \right\|_2^2\right].
\end{equation}

During inference, for a video with $K$ frames where the model processes at most $L$ frames per pass ($K\!\gg\!L$), we first outpaint the selected key frames, then iteratively apply the interpolation model between each neighboring key-frame pair until all frames are outpainted.

\section{Experiments}

\subsection{Experimental Setup}

\noindent\textbf{Implementation Details. }
We initialize the spatial backbone from Stable Diffusion v2.0 inpainting~\cite{sd2inpainting}. We train on 5M videos selected from WebVid-10M~\cite{webvid}. During training, we freeze the spatial pre-trained weights and optimize only the newly introduced temporal and global-guidance modules. From each video, we sample 16 frames, apply center cropping, and resize to $320\times320$. We use a batch size of 32 and AdamW with a learning rate of $2\times10^{-5}$ and 1k-step warm-up. The base model is trained for 670k iterations, and the interpolation model is fine-tuned for 180k iterations.

To synthesize training inputs, we randomly generate masks under three modes: \emph{periphery}, \emph{single-edge}, and \emph{dual-edge} (covering both vertical and horizontal borders). We use a sampling ratio of 0.4/0.1/0.5 for the three modes, respectively. The masked proportion is uniformly sampled from $[0.2,0.8]$ to cover diverse outpainting extents.

\noindent\textbf{Evaluation Datasets and Metrics. }
We evaluate on two widely used benchmarks for video outpainting, DAVIS~\cite{DAVIS2016} and YouTube-VOS~\cite{youtubevos}. Following M3DDM~\cite{M3DDM}, we report PSNR and SSIM~\cite{psnr_ssim}, LPIPS~\cite{lpips}, and FVD~\cite{fvd}.

\subsection{Qualitative Results}

\begin{figure}[!t]
  \centering
  \includegraphics[width=\linewidth]{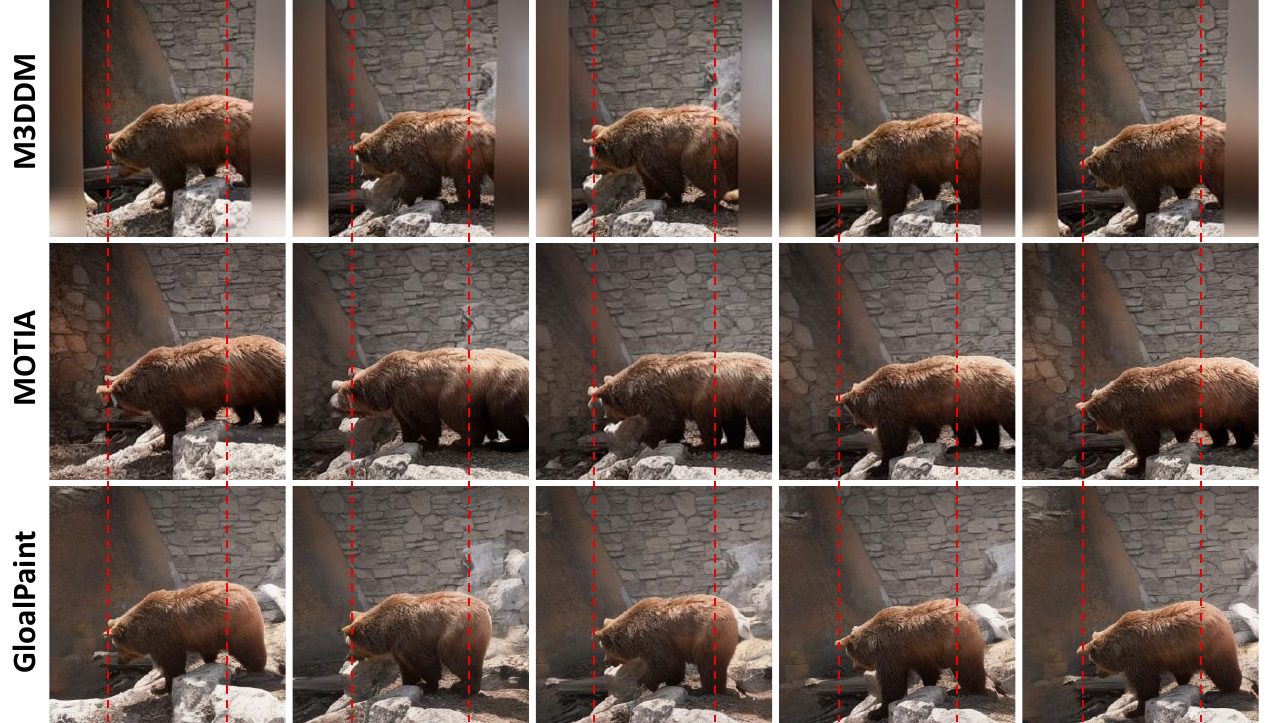}
  \caption{Qualitative comparison with M3DDM~\cite{M3DDM} and MOTIA~\cite{MOTIA}.}
  \label{fig:Qualitative_Comparison}
\end{figure}

Fig.~\ref{fig:Qualitative_Comparison} compares GlobalPaint with M3DDM~\cite{M3DDM} and MOTIA~\cite{MOTIA}. M3DDM often yields over-smoothed and blurry outpainted regions. MOTIA relies on per-sample test-time fine-tuning, leading to higher computation and latency, but it still produces implausible distortions (e.g., duplicated limbs or distorted object shapes). In contrast, GlobalPaint generates sharper and more structurally consistent completions, maintaining better cross-frame coherence without such artifacts. A detailed computational complexity analysis is provided in Appendix~C.

\subsection{Quantitative Results}

\begin{table*}[!t]
\centering
\caption{Quantitative evaluation on DAVIS and YouTube-VOS.}
\label{table:results}
\begin{tabular}{lcccccccc}
\toprule
\multirow{2}{*}{Method} &
\multicolumn{4}{c}{DAVIS} &
\multicolumn{4}{c}{YouTube-VOS} \\
\cmidrule(lr){2-5} \cmidrule(lr){6-9}
& PSNR$\uparrow$ & SSIM$\uparrow$ & LPIPS$\downarrow$ & FVD$\downarrow$
& PSNR$\uparrow$ & SSIM$\uparrow$ & LPIPS$\downarrow$ & FVD$\downarrow$ \\
\midrule
Dehan~\cite{Dehan}   & 17.96 & 0.6272 & 0.2331 & 363.1  & 18.25 & 0.7195 & 0.2278 & 149.7 \\
M3DDM~\cite{M3DDM}   & 20.26 & 0.7082 & 0.2026 & 300.0  & 20.20 & 0.7312 & 0.1854 & 66.62 \\
MOTIA~\cite{MOTIA}   & 20.36 & 0.7578 & 0.1595 & 286.3  & 20.25 & 0.7636 & 0.1727 & \textbf{58.99} \\
GlobalPaint          & \textbf{20.91} & \textbf{0.7621} & \textbf{0.1540} & \textbf{227.8}
                     & \textbf{20.89} & \textbf{0.7938} & \textbf{0.1643} & 60.49 \\
\bottomrule
\end{tabular}
\end{table*}

Following M3DDM~\cite{M3DDM}, we evaluate on the DAVIS 2017 trainval split~\cite{DAVIS2016} and the YouTube-VOS test split~\cite{youtubevos} under two horizontal outpainting ratios (0.25 and 0.666), and report the average over the two settings. As shown in Table~\ref{table:results}, GlobalPaint improves reconstruction quality and yields more natural motion. On DAVIS, it achieves an FVD of 227.8, reducing FVD by 24.1\% and 20.4\% compared to M3DDM~\cite{M3DDM} and MOTIA~\cite{MOTIA}, respectively. On YouTube-VOS, GlobalPaint consistently improves PSNR/SSIM/LPIPS over prior methods and attains a competitive FVD.



\subsection{Ablation Study}
Unless otherwise stated, ablations are conducted on DAVIS 2016 with a horizontal outpainting ratio of 0.5.

\noindent\textbf{Enhanced Spatial-Temporal module.}
Standard 1D temporal adaptation mainly aggregates information along time and has limited capability to model joint spatiotemporal interactions within a local neighborhood. We therefore add $5\times5\times T$ 3D windowed attention after the 1D temporal layers to improve spatiotemporal feature mixing at a feasible cost. As shown in Table~\ref{tab:ablation}, introducing EST reduces FVD from 373.42 to 312.41 (16.3\% relative reduction) compared to the baseline model without window attention. Additional ablations on the windowed attention size are provided in Appendix~B.

\begin{table}[!t]
  \caption{Ablation study on EST module and global feature guidance.}
  \label{tab:ablation}
  \centering
  \begin{tabular}{lcccc}
    \toprule
    Method & PSNR$\uparrow$ & SSIM$\uparrow$ & LPIPS$\downarrow$ & FVD$\downarrow$ \\
    \midrule
    baseline & 19.32 & 0.6895 & 0.1589 & 373.42 \\
    EST      & 19.55 & 0.6966 & 0.1509 & 312.41 \\
    EST+G    & \textbf{20.17} & \textbf{0.7296} & \textbf{0.1501} & \textbf{273.53} \\
    \bottomrule
  \end{tabular}
\end{table}

\begin{figure}[!t]
  \centering
  \includegraphics[width=\linewidth]{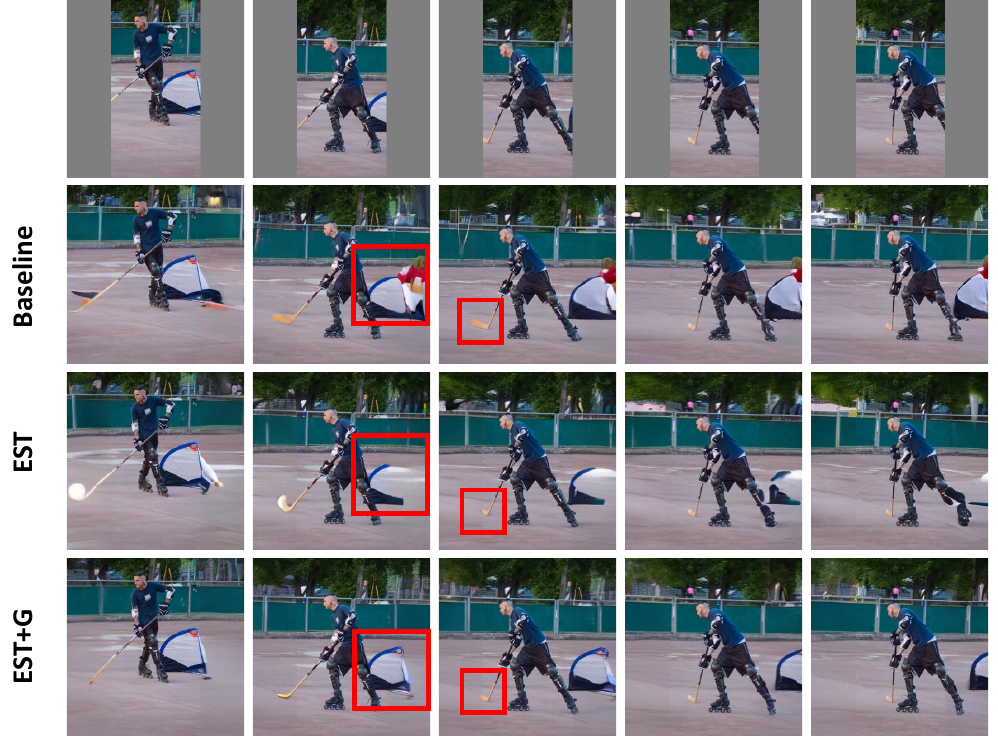}
  \caption{Ablation on EST module and global feature guidance.}
  \label{fig:ablation_est_g}
\end{figure}

\begin{figure*}[!t]
  \centering
  \includegraphics[width=\textwidth]{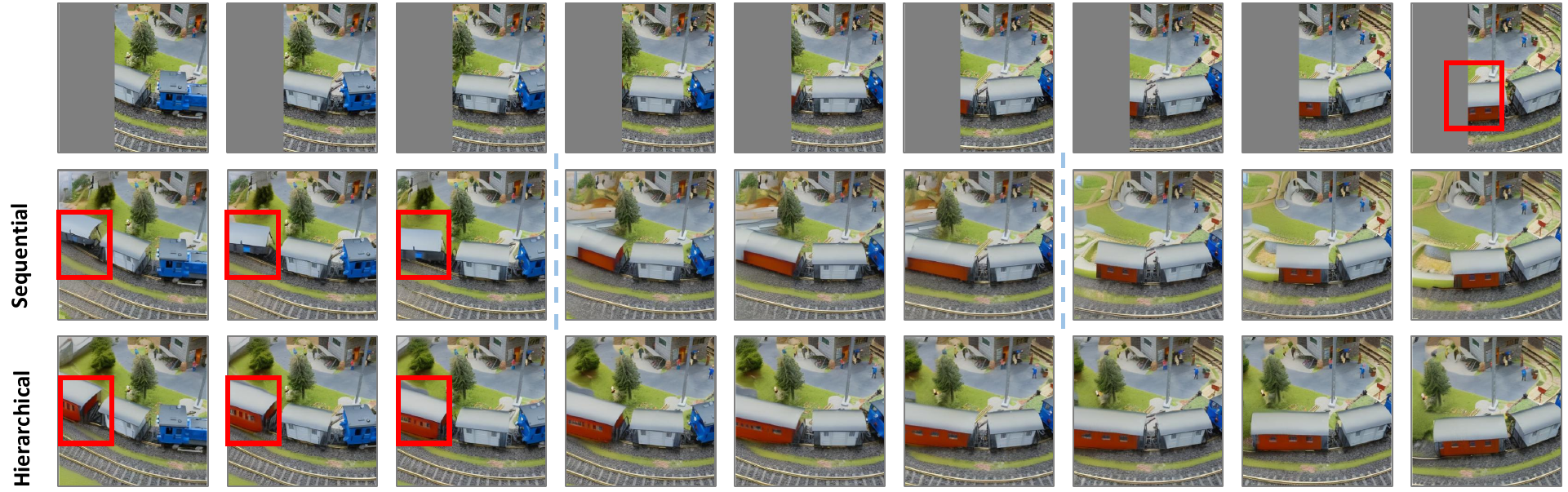}
  \caption{Ablation comparing hierarchical processing and sequential clip-by-clip processing.}
  \label{fig:hierarchical}
\end{figure*}

\noindent\textbf{Global feature guidance.}
When missing-border content appears in other frames, leveraging such observed cues is crucial for maintaining sequence-level consistency. Our global feature extractor aggregates information from observed regions across all frames into compact global tokens and conditions denoising process. Adding global guidance on top of EST (EST+G) further reduces FVD to 273.53 (12.4\% relative reduction over EST), and improves perceptual quality as shown in Fig.~\ref{fig:ablation_est_g}. More ablations and comparisons on the global feature guidance are included in Appendix~B.


\noindent\textbf{Hierarchical processing.}
We compare our hierarchical pipeline with sequential clip-by-clip processing. Key frames sampled at a fixed interval cover a longer temporal span than consecutive short clips, providing richer global context for completing salient objects that gradually enter the view. As illustrated in Fig.~\ref{fig:hierarchical}, sequential processing can produce inconsistent object appearance across clips (e.g., the red carriage), whereas hierarchical processing yields more consistent results.

\section{Conclusion}
In this study, we introduce GlobalPaint, an innovative method aimed at generating spatiotemporal coherent content for video outpainting tasks. Operating within a hierarchical processing framework, our model integrates enhanced spatiotemporal modules and leverages global features to steer the generation process, ensuring global coherence across the entire video sequence. The effectiveness and reliability of GlobalPaint have been rigorously confirmed through a broad array of experiments.

\hypersetup{urlcolor=black}
\bibliographystyle{IEEEbib}
\bibliography{main}

\section*{APPENDIX}

\setcounter{section}{0}
\renewcommand{\thesection}{\Alph{section}}
\renewcommand{\thesubsection}{\Alph{section}.\arabic{subsection}}
\renewcommand{\thesubsubsection}{\Alph{section}.\arabic{subsection}.\arabic{subsubsection}}

\section{Implementation Details}
\noindent\textbf{Global Feature.}
To inject long-range cues with manageable cost, we extract global features from the observed regions of all frames.
Specifically, a frozen OpenCLIP encoder~\cite{openclip} produces context tokens $f_c \in \mathbb{R}^{N\times C}$ with $C{=}1280$.
For a 16-frame clip, we obtain $N{=}16\times257{=}4112$ tokens.
We then apply the Global Feature Extractor (Sec.~3) to compress these context tokens into $M{=}256$ global tokens $f_g \in \mathbb{R}^{M\times C}$, which are injected into the denoising UNet via cross-attention.
This token resampling step reduces redundancy and substantially lowers the attention cost while preserving global information.

\noindent\textbf{Sampling Details.}
We use an Euler EDM sampler~\cite{edm} with 50 steps.
During training, the classifier-free guidance (CFG) scale~\cite{cfg} is set to 7.5.
At inference, we use the video folder name as the prompt for DAVIS.
For YouTube-VOS, which does not provide text descriptions, we generate a caption from the middle frame using GPT-4V~\cite{chatgpt,gpt4-technical-report,gpt4v-system-card,gpt4v-technical-work} and use it as the prompt for the entire sequence.
Following the evaluation settings of M3DDM~\cite{M3DDM}, we set CFG to 5.0 for the 0.25 mask ratio and 2.0 for the 0.666 mask ratio, and keep the sampler configuration unchanged.

\noindent\textbf{Benchmark Details.}
We evaluate on DAVIS~\cite{DAVIS2016,davis2017} and YouTube-VOS~\cite{youtubevos}, which are widely used for benchmarking video outpainting despite being originally introduced for video object segmentation.
We report main results on the DAVIS 2017 trainval split (90 videos) and the YouTube-VOS test split (508 videos), and use DAVIS 2016 (50 videos) for ablation studies.

\section{Additional Experimental Results}

\noindent\textbf{Ablation on windowed attention size.}
We study the effect of the window size used in the 3D windowed attention of the EST module.
As shown in Table~\ref{tab:window_size}, a $5\times5\times T$ window (EST-55T) outperforms a $3\times3\times T$ window (EST-33T) after 40k training iterations.
Under our default setting, the smallest UNet feature map is $5{\times}5$, so a $5{\times}5$ window effectively performs full attention at this stage, improving global context aggregation while keeping computation tractable; windows larger than $5{\times}5$ rapidly increase the attention cost and can become prohibitively expensive.

\begin{table}[!t]
  \caption{Ablation on window size.}
  \label{tab:window_size}
  \centering
  \small
  \setlength{\tabcolsep}{4pt}
  \begin{tabular}{lcccc}
    \toprule
    Method & PSNR$\uparrow$ & SSIM$\uparrow$ & LPIPS$\downarrow$ & FVD$\downarrow$ \\
    \midrule
    EST-55T & \textbf{18.99} & \textbf{0.6778} & \textbf{0.1902} & \textbf{528.48} \\
    EST-33T & 18.81 & 0.6731 & 0.1928 & 559.22 \\
    \bottomrule
  \end{tabular}
\end{table}

\noindent\textbf{Effect of the number of learnable global tokens.}
We vary the number of learnable query tokens in the Global Feature Extractor to balance representational capacity and optimization stability.
Table~\ref{tab:global_tokens} shows results at 40k iterations.
Using too few tokens ($M{=}64$) over-compresses global context and degrades perceptual quality (LPIPS), whereas too many tokens ($M{=}1024$) makes optimization harder and worsens motion quality (FVD).
We use $M{=}256$ by default.

\begin{table}[!t]
  \caption{Ablation on number of global tokens.}
  \label{tab:global_tokens}
  \centering
  \small
  \setlength{\tabcolsep}{4pt}
  \begin{tabular}{lcccc}
    \toprule
    Method & PSNR$\uparrow$ & SSIM$\uparrow$ & LPIPS$\downarrow$ & FVD$\downarrow$ \\
    \midrule
    GlobalPaint-64   & \textbf{19.10} & \textbf{0.6792} & 0.1906 & 477.12 \\
    GlobalPaint-256  & 19.05 & 0.6777 & 0.1899 & \textbf{474.59} \\
    GlobalPaint-1024 & 19.01 & 0.6753 & \textbf{0.1883} & 513.14 \\
    \bottomrule
  \end{tabular}
\end{table}

\noindent\textbf{Analysis of global feature extraction.}
We extract OpenCLIP features from the observed regions of all frames using the penultimate-layer tokens (including the class token), rather than using only a single pooled token.
We compare our design with the lightweight encoder in M3DDM~\cite{M3DDM} (denoted as Lightweight-Enc), which provides global features using a lightweight video encoder.
As shown in Table~\ref{tab:ablation_global}, Lightweight-Enc attains similar PSNR/SSIM/LPIPS but increases FVD from 474.59 to 551.28 (a 16.2\% relative increase), indicating weaker temporal realism.

\begin{table}[!t]
  \caption{Ablation on global feature design.}
  \label{tab:ablation_global}
  \centering
  \small
  \setlength{\tabcolsep}{4pt}
  \begin{tabular}{lcccc}
    \toprule
    Method & PSNR$\uparrow$ & SSIM$\uparrow$ & LPIPS$\downarrow$ & FVD$\downarrow$ \\
    \midrule
    GlobalPaint     & 19.05 & 0.6777 & 0.1899 & \textbf{474.59} \\
    Lightweight-Enc & \textbf{19.10} & 0.6762 & 0.1900 & 551.28 \\
    \bottomrule
  \end{tabular}
\end{table}

\section{Computational Complexity}
We summarize trainable parameters (Par.), computational cost (FLOPs), peak inference GPU memory (I-GPU), and end-to-end inference time (I-Time) in Table~\ref{tab:compute}. 
GlobalPaint freezes the pretrained 2D spatial backbone and trains only the newly added temporal and global modules, resulting in fewer trainable parameters than M3DDM~\cite{M3DDM}. 
Under the same setting (one A100 GPU, 16 frames at $320{\times}320$), GlobalPaint also requires fewer FLOPs and substantially less inference time while using comparable memory. 
This improvement is partly because M3DDM~\cite{M3DDM} adopts a specialized classifier-free guidance scheme that increases compute overhead.

We further include MOTIA~\cite{MOTIA}, whose inference is dominated by per-video test-time adaptation: it fine-tunes on each test sample for 800 steps (243\,s) and then performs pattern-aware outpaint (16\,s), leading to a total of 259\,s inference time per sample, which is considerably slower than feed-forward methods that do not require test-time optimization.

\begin{table}[!t]
  \caption{Analysis on model parameters and computation.}
  \label{tab:compute}
  \centering
  \begin{tabular}{lccccc}
    \toprule
    Method & Par.\,(M) & FLOPs\,(G) & I-GPU\,(GB) & I-Time\,(s) \\
    \midrule
    M3DDM~\cite{M3DDM} & 1299 & 15667 & 30 & 44  \\
    MOTIA~\cite{MOTIA} & 7.49 & -- & 13 & 259 \\
    GlobalPaint        & 1024 & 11177 & 31 & \textbf{17}  \\
    \bottomrule
  \end{tabular}
\end{table}

\begin{figure}[h]
  \centering
  \includegraphics[width=\linewidth]{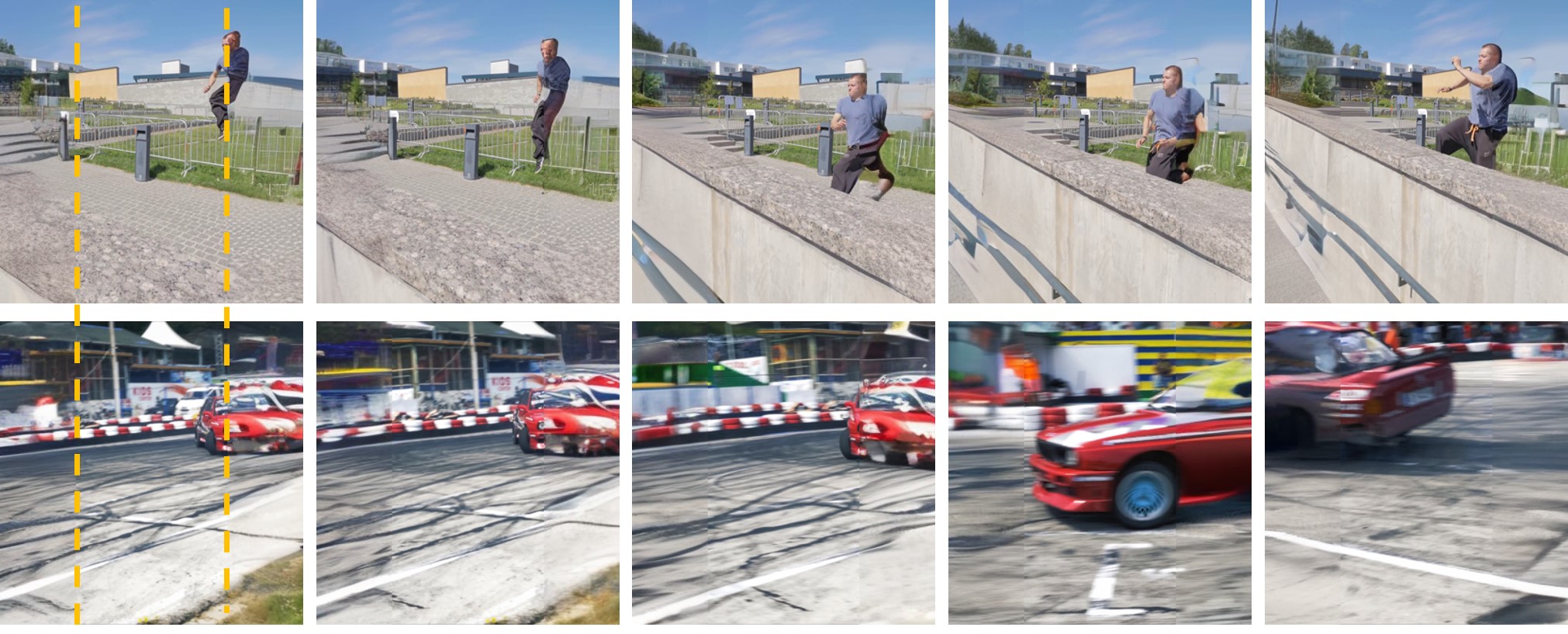}
  \caption{Bad cases in our model. Contents outside the yellow lines are outpainted by our model.}
  \label{fig:limit}
\end{figure}

\section{Limitation and Future Work}
Fig.~\ref{fig:limit} presents examples where our model fails to generate high-quality results. Under extreme dynamics such as rapid motion and severe camera shake (e.g., parkour), the model may yield inaccurate limb completion, and similar degradations occur for fast-moving vehicles when the \emph{input video} contains strong motion blur and lost details. These issues are likely due to limited training coverage of heavy-shake/extreme-motion scenarios and the inherent ambiguity introduced by blurred or low-detail inputs; future work will improve robustness by enriching such data with targeted augmentations and by exploring motion-aware conditioning (e.g., stronger temporal cues or deblurring-aware features) to better leverage uncertain observations.

\section{More Comparison} 
\begin{figure*}[h]
  \centering
  \includegraphics[width=\textwidth]{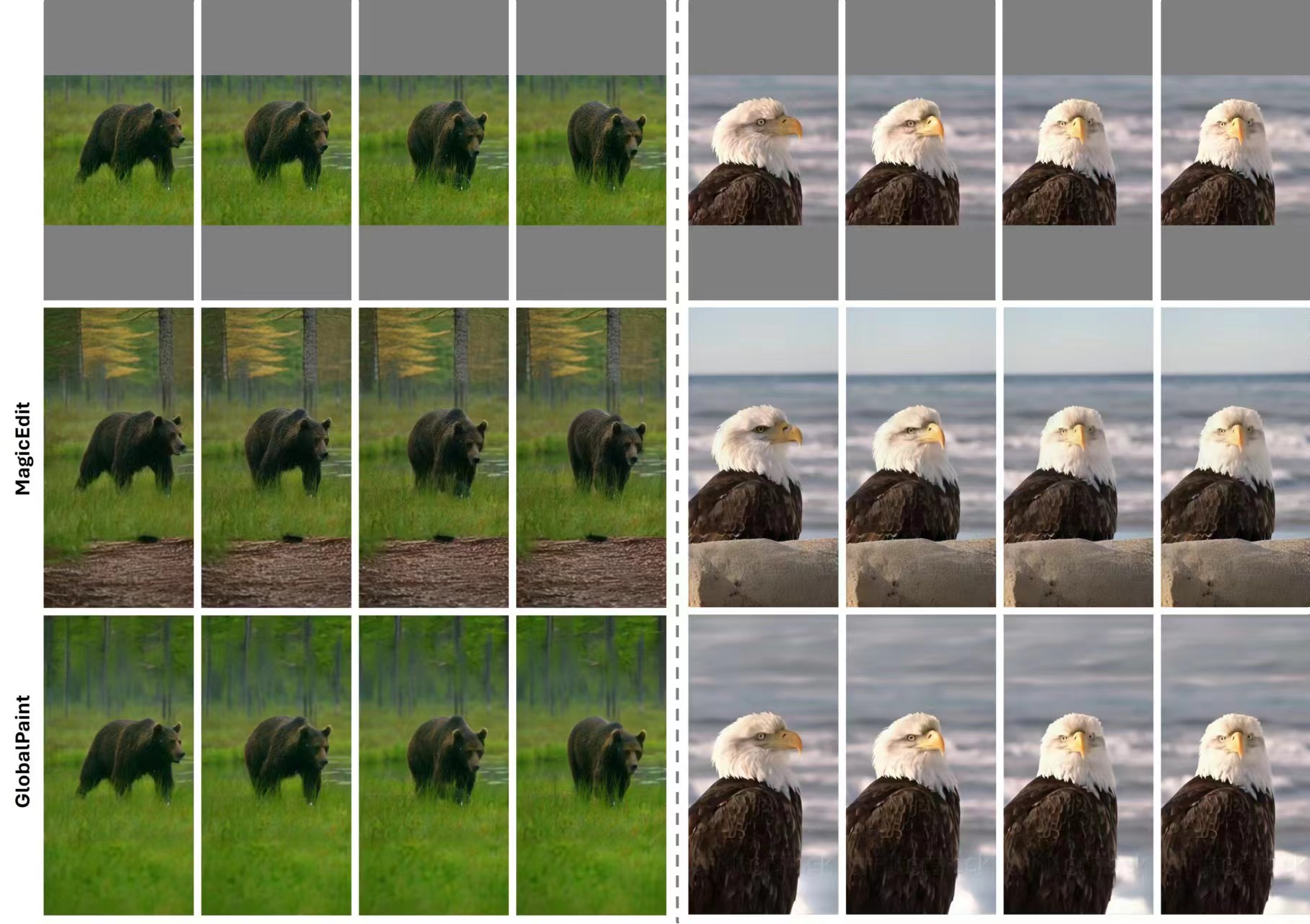}
  \caption{Comparsion between GlobalPaint and MagicEdit \cite{liew2023magicedit}.}
  \label{fig:compare magicedit}
\end{figure*}

In Fig.~\ref{fig:compare magicedit}, we compare the results of video outpainting from an aspect ratio of 1:1 to 2:1 using GlobalPaint and MagicEdit \cite{liew2023magicedit}. As MagicEdit has not publicly released their code, our comparison is based on a few cherry-picked examples from MagicEdit website. Our model achieves comparable performance on these selected examples from MagicEdit. Limited by a 20MB size restriction for supplementary materials, we have showcased only two examples. Nonetheless, our model also performs well on other examples.

\begin{figure*}[h]
  \centering
  \includegraphics[width=\textwidth]{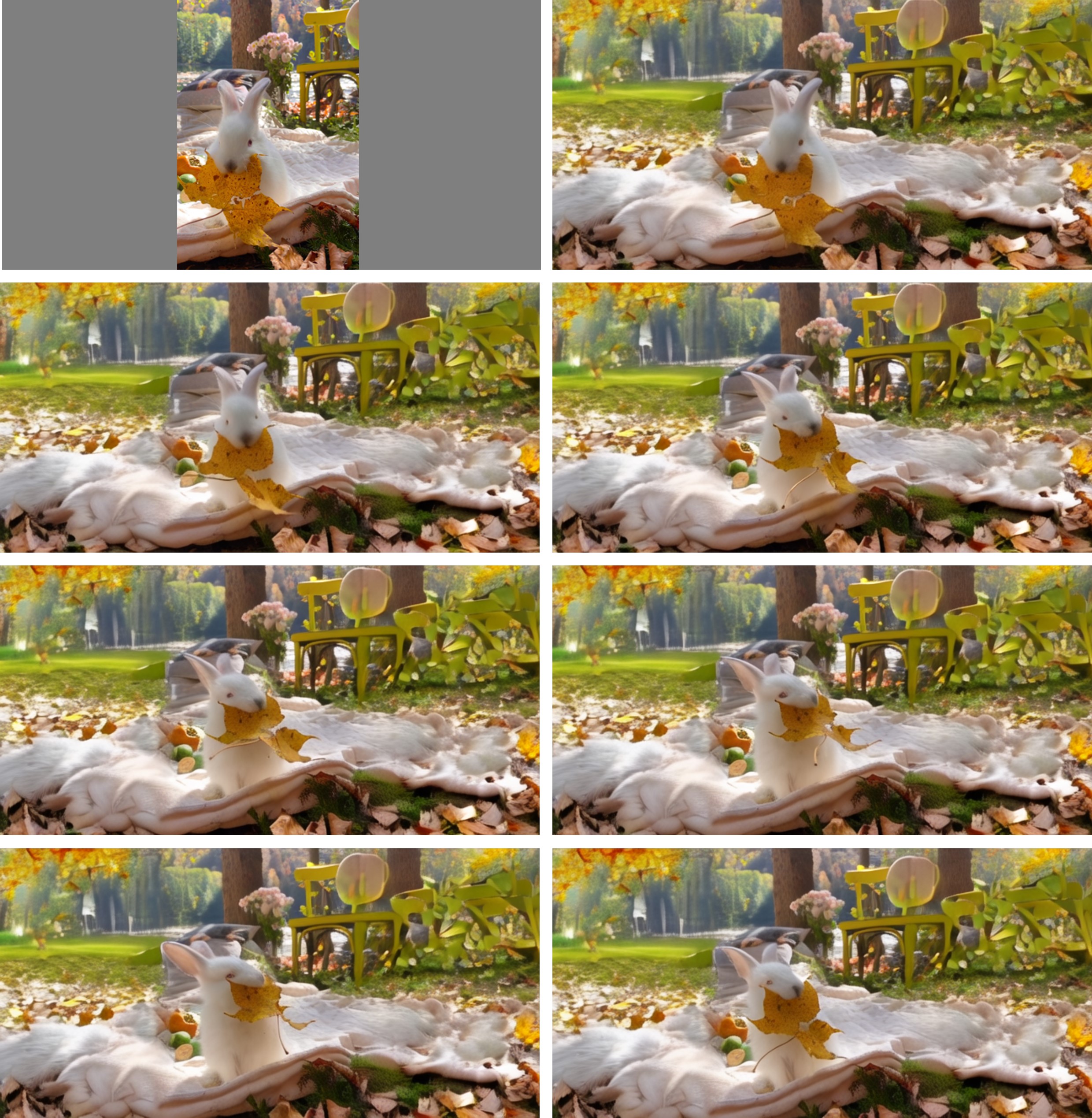}
  \caption{Result of GlobalPaint outpainting a video from $320 \times 216$ to $320 \times 640$.}
  \label{fig:rabbit640}
\end{figure*}

\begin{figure*}[h]
  \centering
  \includegraphics[width=\textwidth]{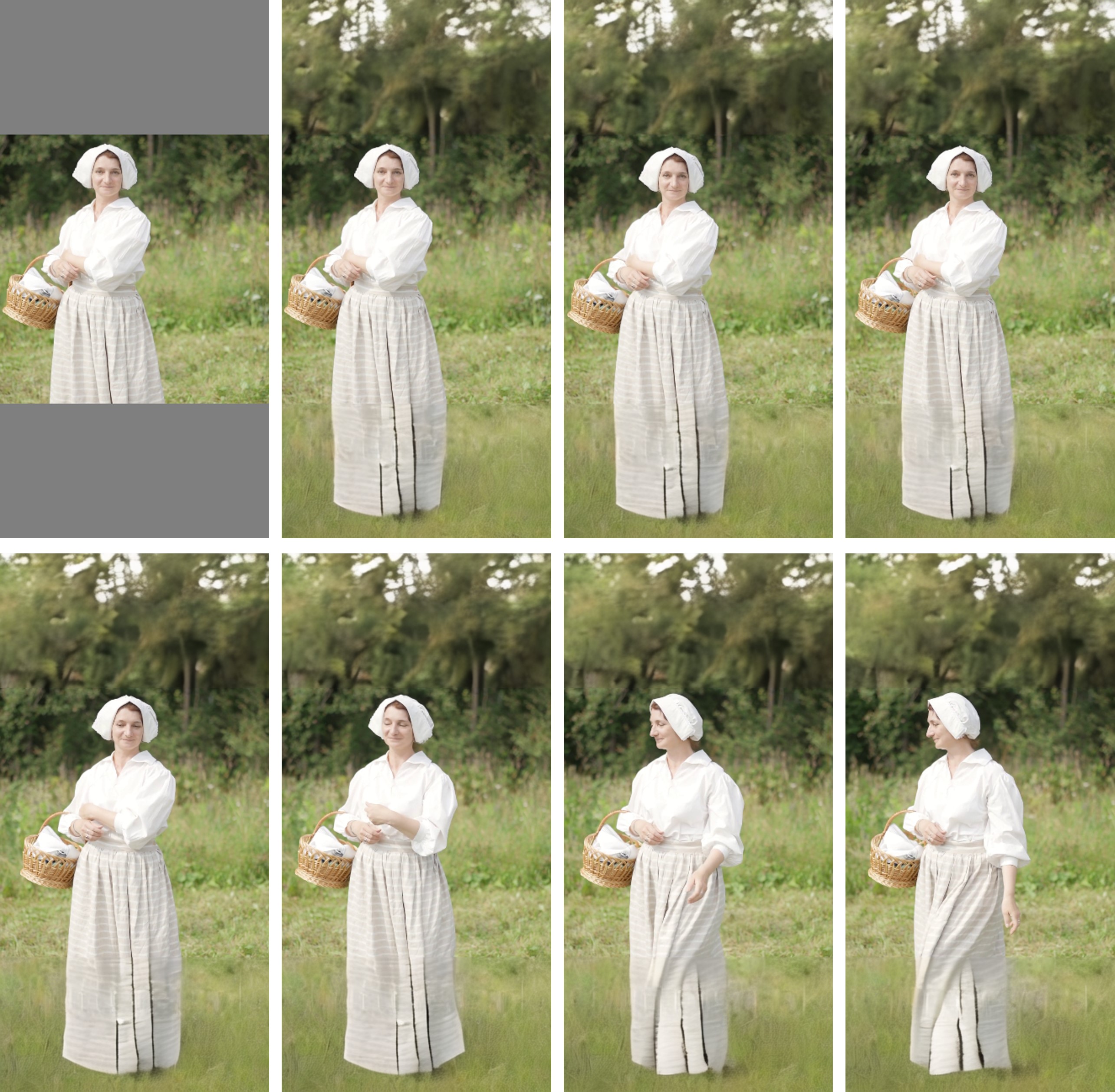}
  \caption{Result of GlobalPaint outpainting a video from $320 \times 320$ to $640 \times 320$.}
  \label{fig:woman640}
\end{figure*}

\section{More Visualizations}
Although GlobalPaint is trained at a resolution of $320{\times}320$, it generalizes well to non-square aspect ratios at test time. As shown in Fig.~\ref{fig:rabbit640} and Fig.~\ref{fig:woman640}, we evaluate the model at $640{\times}320$ and $320{\times}640$, respectively, and obtain plausible outpainting results, demonstrating strong adaptability to varying aspect ratios.

When increasing the output resolution, we observe occasional temporal flickering in fine-grained textures, which we attribute to the limited temporal modeling of the 2D VAE decoder. To alleviate this issue, we augment the VAE decoder with lightweight temporal layers following prior work~\cite{videoLDM}, which effectively reduces high-frequency flicker in the generated videos.

\end{document}